\documentclass[letterpaper]{article} %
\usepackage{aaai24}  %
\usepackage{times}  %
\usepackage{helvet}  %
\usepackage{courier}  %
\usepackage[hyphens]{url}  %
\usepackage{graphicx} %
\usepackage{natbib}  %
\usepackage{caption} %
\usepackage{algorithm}
\usepackage{algpseudocode}

\usepackage{amsmath}
\usepackage{amssymb}
\usepackage{colortbl}
\usepackage{xcolor}

\usepackage{newfloat}
\usepackage{listings}
\DeclareCaptionStyle{ruled}{labelfont=normalfont,labelsep=colon,strut=off} %
\floatstyle{ruled}
\newfloat{listing}{tb}{lst}{}
\floatname{listing}{Listing}
\title{Learning Communication Policies for Different Follower Behaviors \\ in a Collaborative Reference Game}

\author {
    Philipp Sadler\textsuperscript{\rm 1}{\rm ,} 
    Sherzod Hakimov\textsuperscript{\rm 1}{\rm and} 
    David Schlangen\textsuperscript{\rm 1,2}
}
\affiliations {
    \textsuperscript{\rm 1}CoLabPotsdam / Computational Linguistics\\
    Department of Linguistics, University of Potsdam, Germany\\
    \textsuperscript{\rm 2}German Research Center for Artificial Intelligence (DFKI), Berlin, Germany\\
    firstname.lastname@uni-potsdam.de
}

\begin{document}

\maketitle

\begin{abstract}
Albrecht and Stone (2018) state that modeling of changing behaviors remains an open problem ``due to the essentially unconstrained nature of what other agents may do''. In this work we evaluate the adaptability of neural artificial agents towards assumed partner behaviors in a collaborative reference game. In this game success is achieved when a knowledgeable Guide can verbally lead a Follower to the selection of a specific puzzle piece among several distractors. We frame this language grounding and coordination task as a reinforcement learning problem and measure to which extent a common reinforcement training algorithm (PPO) is able to produce neural agents (the Guides) that perform well with various heuristic Follower behaviors that vary along the dimensions of confidence and autonomy. We experiment with a learning signal that in addition to the goal condition also respects an assumed communicative effort. Our results indicate that this novel ingredient leads to communicative strategies that are less verbose (staying silent in some of the steps) and that with respect to that the Guide's strategies indeed adapt to the partner's level of confidence and autonomy.
\end{abstract}

\begin{figure}[t]
    \begin{center}
        \includegraphics[width=0.35\textwidth]{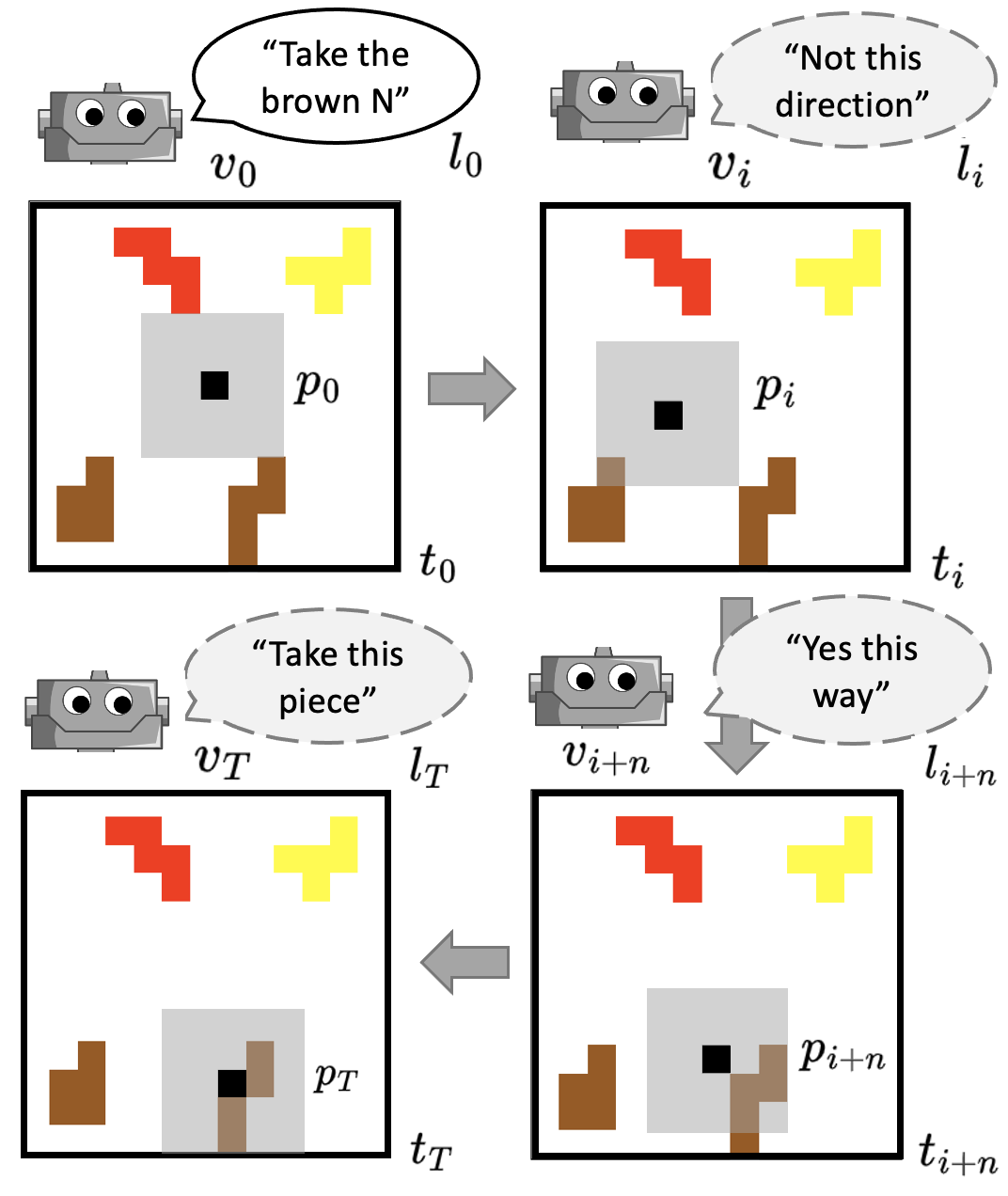}
    \end{center}
    \caption{An exemplary interaction between a Guide and a Follower that controls the gripper (the black dot). The Guide observes the scene $v_0$ and refers to a piece initially with $l_0$. The Follower has only a partial view $p_0$ (the grey box) and might go wrong. The Guide can provide further information based on the Follower's actions until a piece is selected at time step $T$. The Guide should learn that less utterances are necessary with a more autonomous and confident Follower.
    }
    \label{fig:example_board}
\end{figure}

\section{Introduction} %

Sometimes we feel like we could continue another person's sentence. This happens in particular with people we know well or we often interact with. A common phrase coined to this phenomenon is that ``people are on the same wavelength''. And indeed \citet{davidesco_temporal_2023} found that brain activities somewhat synchronize between teachers and students during lessons. Even more surprising, synchronicity becomes a good predictor of the learning success of the students. 
A psycho-linguistic study by \citet{clark_referring_1986} observed the language use of collaborative partners during an ongoing goal-oriented interaction: They (implicitly) agree on newly introduced noun phrases and a common strategy to achieve the goal together. Interestingly, the number of used words drastically decrease during the collaboration. The participants strive towards reduced individual %
efforts while the number of successful outcomes stays high.
We see that human interaction is characterized by synchronicity (adaption) and the reduction of the individual 
effort. In this work we study how (artificial) learning agents adapt to an assumed partner's behavior. For this, we propose a simple, but still challenging vision and language grounding task where two players have to coordinate on the selection of a puzzle piece (Pentomino; \citet{golomb_1996}) while (i) the actual target piece is only known to one of them (the Guide), and (ii) only the other can perform the selection (the Follower). See Figure~\ref{fig:example_board} for an example illustration of this goal-oriented collaborative game.
\citet{clark_using_1996} points out that in these situations language acts as a device for solving the coordination problem: If the participants agree on a mutually desired outcome (the goal; for example taking a specific piece) then their individual participatory actions take part in a joint action. The regularity in behavior, common ground, and the recurrence of the coordination problem lets them settle on conventions -- and ultimately adapt to each other. We think that such capabilities would be essential for future assisting agents that might take part in society someday (see \citet{park2023generative} for a toy example).
\citet{albrecht_autonomous_2018} found that modeling of changing behaviors (or different others) remains an open problem ``due to the essentially unconstrained nature of what other agents may do''. 
Are neural agents capable to adapt to their interactants and converging to strategies that become useful only during the dynamic interaction itself (when the partner's behavior becomes apparent)? 

In this paper, we frame the language coordination and grounding task as a reinforcement learning problem \citep{Sutton1998} and evaluate, if and to which extent a common training algorithm Proximal Policy Optimization (PPO) \citep{schulman_proximal_2017} is able to produce guiding neural agents that perform well with a variety of Follower behaviors in a collaborative setting where the Guide's utterances become language actions. In this scenario, an agent (or possibly multiple ones) take step-wise actions in an observable and dynamic environment to maximize a reward signal. 

The main idea is that we assume an ongoing interaction in which the Follower's behavior changes. After some time the Follower should become more autonomous and more confident in choosing actions and executing its own plan (as pointed out by \citet{clark_referring_1986}). But instead of treating this as a multi-agent setting directly, we follow \citet{DBLP:conf/nips/Yang0HZZL22} (with the notion of assigning different agents to different sub-tasks) and learn separate Guides for each of the (hand-crafted) Follower behaviors (sub-tasks). The resulting policies represent a Guide's communicative strategy at certain points in time of the assumed ongoing interaction.
Our expectations on the learned communicative strategies of the Guide are that in the beginning (with a less autonomous, less confident Follower) more is to be said. And that with a more autonomous and confident Follower the Guide learns that it ``does not need to say anything'' to be successful (and consequently reducing the effort). Our contributions are the following\footnote{Source code is publicly available under: \url{https://github.com/clp-research/different-follower-behaviors}}:

\begin{itemize}
    \item We propose a challenging RL environment: a reference game in which a neural agent (the Guide) has to learn communication strategies that are \textbf{successful and reduce an assumed effort}, and
    \item contribute a plausible Follower policy (the training partner) that is variable on two dimensions: \textbf{confidence} and \textbf{autonomy}, and
   \item present strong baseline Guide policies for this difficult cooperative reference game that are indeed able to balance out episode success and their individual effort by \textbf{learning to stay silent}.
\end{itemize}

\section{Related Work} %

\paragraph{Vision and language navigation.}

The use of natural language to guide an instruction following agent has been heavily studied for the vision and language navigation task \citep{gu_vision-and-language_2022,DBLP:conf/cvpr/NguyenDBD19,DBLP:conf/emnlp/NguyenD19,DBLP:conf/nips/FriedHCRAMBSKD18,DBLP:journals/corr/abs-1907-04957}. For example, \citet{DBLP:conf/emnlp/NguyenD19} train an instruction giver (IG) on a pre-collected dataset of instructions. The Follower is then allowed to ask the IG for more information during task execution. Although the setting is very similar, we distinguish from these works as our Guide has to learn itself when to provide more information to the Follower. In our setting, the language back-channel for the Follower is cut, so that the Guide's timing and utterance choice becomes essential.

\paragraph{Natural language goals in RL.} 

Using natural language to describe the goal state in an RL problem has become a common theme \citep{chevalier-boisvert_babyai_2019,gao_dialfred_2022,padmakumar_teach_2022,pashevich_episodic_2021,suhr_continual_2022}.  This research direction is interesting because it could allow humans to interact more easily with learned agents. There is work that shows that intermediate language inputs are a valuable signal in task-oriented visual environments \citep{co-reyes_guiding_2019,mu_improving_2022}. Indeed \citet{huang_reminder_2023} found that natural language can ``provide a gradient'' towards the goal state. But they also point out the ``brittleness'' of these signals because the language input might align badly with sub-trajectories. A key challenge here is the variability of expressions in language that can be produced and understood in the defined action space. Even in relatively simple environments, there might arise an overwhelming amount of situations for an agent to handle \citep{chevalier-boisvert_babyai_2019}. We weaken the action space exploration problem by using ideas from natural language understanding  \citep{moon_situated_2020,e_novel_2019} and let the guide produce language actions in a well-defined reduced ``intent space''. These intents are then verbalized (using templates; which could be a conditioned pre-trained language model) and given to the follower.

\paragraph{Interactive sub-goal generation in RL.}

\citet{sun_adaplanner_2023} use a pre-trained large language model to generate possible plans (in the form of source code) for the completion of a task. They introduce a distinction between implicit and explicit closed-loop systems that are able to either refine single actions or an entire plan respectively. 
Indeed neural agents perform better when they self-predict sub-goals to be achieved (with an intrinsic reward) instead of reaching for the final goal immediately \citep{jurgenson_goal-conditioned_2023,chane-sane_goal-conditioned_2021,pertsch_long-horizon_2020,jeon_maser_2022}. For example, \citet{lee_controllable_2023} study the task of finding the best route in a simple visual domain by training a sub-goal system that predicts intermediate coordinates. In contrast to them, our guiding agent has to produce a natural language utterance to describe a sub-goal (and we use referring expressions or directions). \citet{gurtler_hierarchical_2021} also address the question of ``when to provide sub-goals'', which is necessary in our task. Nevertheless, in distinction to these works, we treat the sub-goal generation not just as additional information for the follower's success but are interested in the learned communicative strategies themselves. We treat the sub-goal providing guide as an individual participant in the environment similar to a multi-agent setting.

\begin{figure*}[t]
    \begin{center}
        \includegraphics[width=0.95\textwidth]{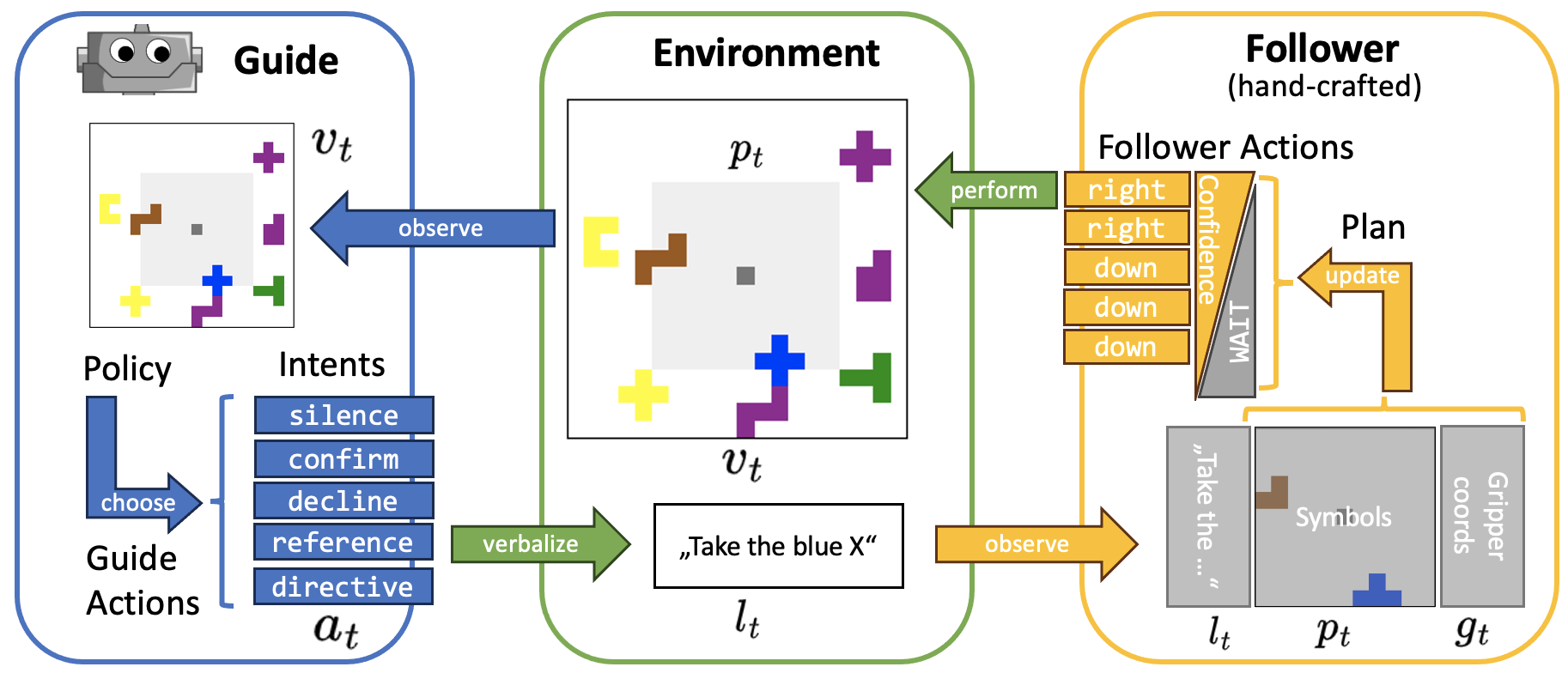}
    \end{center}
    \caption{The general information and decision-making flow of the reference game. The Guide observes $v_t$ which contains the full scene in pixel space and additionally  the gripper position (4th-channel) and target piece (5th-channel). Given this, the Guide chooses an intent action $a_t$ that gets verbalized into a natural language sentence $l_t$. %
    Then, the Follower receives the utterance $l_t$, the gripper coordinate $g_t$ and a symbolic representation of a partial view of the scene $p_t$. The hand-crafted policy updates the plan accordingly based on its given representation of the world. Finally, the Follower's next planned action (or \texttt{wait}) is performed with a certain chance defined by the attached confidence. The process repeats until a piece is taken or time runs out.
    }
    \label{fig:environment}
\end{figure*}

\paragraph{Skill learning in cooperative multi-agent RL.}

We treat both guide and follower as agents in a cooperative setting
and follow work that uses hand-crafted policies \citep{wang_influencing_2022,ghosh_towards_2020,xie_learning_2021} (here a follower that is able to mimic behavior that varies in autonomy and confidence). In this sense, our approach is similar to heterogeneous skill learning \citep{chang_e-mapp_nodate,liu_heterogeneous_nodate,hu_enabling_2023} where a single agent is trained to acquire a variety of skills (in our case communicative strategies). This is, in particular, helpful due to the differences in the action spaces of the guide (language acts) and the follower (movements). In addition, this method (of having a fixed hand-crafted follower policy) allows us to avoid the problem of emergent communication where agents agree on a language that becomes inaccessible by humans \citep{lowe_pitfalls_2019,mul_mastering_2019,kolb_learning_2019}.

\section{The CoGRIP-GL Reference Game} %

We use the \textbf{Co}llaborative \textbf{G}ame of \textbf{R}eferential and \textbf{I}nteractive language with \textbf{P}entomino pieces (CoGRIP) (reference suppressed) %
and extend it for \textbf{G}uidance \textbf{L}earning (CoGRIP-GL). A Guide uses natural language to instruct a Follower to select a specific target piece using a gripper. In this setting, both players are constrained as follows: The Guide can provide utterances but cannot move the gripper. The Follower can move the gripper but is not allowed to provide an utterance. This asymmetry in knowledge and skill forces them to work together and coordinate. \citet{zarries_pentoref_2016} found that such a reference game leads to diverse language use on the Guide's side. The most similar environment is from \citet{mordatch2017emergence} who studied cooperative communication where a listener has to navigate to one of three landmarks. The target is only known by a speaker that can not move. The speaker has to learn how to make use of a restricted vocabulary based on a dense reward signal (the listener's distance to the ground-truth landmark). In our game, we only provide a sparse reward and the communication signals become verbalized into language utterances.

\subsection{Problem Formulation}
The Guide has to provide utterances that are useful for the Follower to navigate and select the correct target piece. We frame this task as an RL problem with sparse rewards. At each time-step $t$, given an observation $o_t \in \mathcal{O}$ of the environment, the Guide has to choose an action $a_t$ such that the overall resulting sequence of actions $(a_0,...,a_t,...,a_T)$ (which become verbalized into $(l_0,...,l_t,...,l_T)$) maximizes the sparse reward $\mathcal{R}(o_T)=r$ that is given on episode end when a piece is selected by the Follower or $t$ reaches $T_{max}=30$. This maximal number of steps is sufficient to navigate to the target piece with some room for error on our $21\times21$ tile maps as the Follower is always starting in the center of the map (the farthest tile would be 10 horizontal plus 10 vertical steps away) and allows quicker training.

\subsection{Actions}

We let the Guide predict ``intent'' actions and translate them into sentences instead of predicting words directly to reduce the agent's burden on action space exploration (later this verbalization process could be done by a language generation system). Here we focus on the Guide's choice among five intent categories:  \texttt{silence, confirm, decline, directive, reference} (Figure~\ref{fig:environment}). 
For the \texttt{directive}s, we allow more fine-grained control over the utterance production, so that the agent has to choose between \texttt{left, right, up, down} and \texttt{take}. And similarly, for the \texttt{reference}s this means that the agent has to choose among possible preference orders \texttt{PCS, PSC, SPC, CPS, SCP} and \texttt{CSP}, in which \texttt{P}, \texttt{C} and \texttt{S} stand for piece, color, and shape, respectively. These preference orders (PO) define the order in which properties are compared between the target piece and its distractors. This means, for example, that a \texttt{CSP}-based reference is likely to mention the target piece's color because the color is tried first to distinguish the target from its distractors (and it is very unlikely that all pieces share the same color).
These six \texttt{reference} actions, five \texttt{directive} actions, \texttt{silence}, \texttt{confirm} and \texttt{decline} lead to a total of $|A|=14$ actions. In comparison, the vocabulary contains 37 tokens and the maximal sentence length is 12 which results in $37^{12}$ possible utterances when predicting individual words instead of intents.

\subsection{Verbalization}

The chosen intent is then verbalized based on templates by application of the following rules:

\begin{itemize}
    \item \texttt{silence} $\rightarrow$ \texttt{<empty string>}
    \item \texttt{confirm} $\rightarrow$ \texttt{Yes this [way|<piece>]}
    \item \texttt{decline} $\rightarrow$ \texttt{Not this [way|<piece>]}
    \item \texttt{directive(\texttt{take})} $\rightarrow$ \texttt{Take <piece>}
    \item \texttt{directive}(\texttt{dir}) $\rightarrow$ \texttt{Go <\texttt{dir}>}
    \item \texttt{reference(PO)} $\rightarrow$ \texttt{Take the <\textsc{IA}(PO)>}
\end{itemize}

where \texttt{<piece>} resolves to a piece's color and shape when the current gripper position is located over a piece (or otherwise simply \texttt{piece}). The direction \texttt{<dir>} resolve to the according intent name. 
The fine-grained reference intent (\texttt{PO}) is given to the ``Incremental Algorithm'' \citep{dale_computational_1995}, which produces the referring expression for reference verbalization (Appendix~\ref{appendix:environment}). 

\subsection{Rewards}

Following \citet{chevalier-boisvert_babyai_2019}, we define a basic sparse reward for playing the game:

\begin{equation}
 \mathcal{R_{\text{Game}}} = 1 - 0.9 * (T / T_{\text{max}})   
\end{equation}

In addition, we introduce a sparse reward for the Guide's individual effort in an episode:

\begin{equation}
 \mathcal{R_{\text{Guide}}} = 1 - 0.9 * (E_{\text{Guide}} / T_{\text{max}})   
\end{equation}

where the Guide's effort $E_{\text{Guide}}$ is the sum over the assumed efforts of taking the respective actions:
\begin{equation}
 E_{\text{Guide}} = \sum_{t=1}^T\begin{cases}
 0, & \text{if $a_t$ $\in$ \{\texttt{silence}\}} \\ 
 1.0, & \text{if $a_t$ $\in$ \{\texttt{confirm,decline}\}}\\
 1.1, & \text{if $a_t$ $\in$ \{\texttt{directive}\}}\\ 
 1.2, & \text{if $a_t$ $\in$ \{\texttt{reference}\}}
 \end{cases}
\end{equation}

These action-based efforts follow the assumed cognitive load for producing them i.e.\ saying nothing is the cheapest and comparing pieces with each other to produce a reference is the highest.
Finally, we give an additional reward ($\mathcal{R}_\text{Outcome}$) of $+1$ when the correct piece or a penalty of $-1$ if the wrong or no piece has been taken at all, so that:
\begin{equation}
 \mathcal{R} = 
     (\mathcal{R_{\text{Game}}} + \mathcal{R_{\text{Guide}}}) / 2 + \mathcal{R_{\text{Outcome}}} 
\end{equation}

Given this formulation, the Guide has to play the game by being active (not just stay silent), achieve the goal (get the bonus) and reduce its individual effort (stay mostly silent) to reach a high reward.

\subsection{Observations}

The environment exposes at each time-step $t$ an observation $o_t$ that contains the following:
\begin{itemize}
    \item the Follower's gripper coordinates $g_t = (x,y)$
    \item the Guide's utterance $l_t$ (might be empty) 
    \item a full view of the scene $v_t$ for the Guide 
    \item a partial view $p_t$ of the scene for the Follower
\end{itemize}

The visual observations are represented as 2-dimensional images (with RGB color channels), but the Follower only receives a $11\times11$-sized cut out, centered on the gripper position (see Figure~\ref{fig:environment}). 
We add a 4th channel to the visual observations to indicate the gripper position by setting the values to zero at $g_t$ and one otherwise. In addition, the Guide is informed about the target piece coordinates by setting the according values to zero for the target piece and ones otherwise on the 5th channel of its visual observation. A piece occupies five adjacent tiles and is not allowed to overlap with another one. For our purposes, the Follower receives a symbolic representation of the partial view (as a neural learner might receive) where color and shapes are directly represented as numbers (and not pixels; see Appendix~\ref{appendix:environment}).

\begin{table}[t]
\centering
\begin{tabular}{llll}
\hline
\textbf{}  & \textbf{TPS} & \textbf{Tasks} & \textbf{Boards} \\ \hline
Training   & 275          & 2500           & 700             \\
Validation & 25           & 175            & 175             \\
Testing    & 60           & 420            & 420             \\ \hline
\end{tabular}
\caption{The number of tasks and boards in each data split. The target pieces for the tasks are chosen from non-overlapping sub-sets of target piece symbols (TPS). For evaluation splits, we mix-in training pieces as distractors. We construct boards with up to 7 distractors (and at least 1).
}
\label{table:tasks}
\end{table}

\subsection{Tasks}

The task is that a Guide provides utterances to a Follower that has to take an intended target piece among several other pieces (the distractors). Thus, a game instance of this task is defined by (i) the number and identity of pieces on the board, (ii) including which of these is the the target piece, (iii) and by the size of the board (see Figure~\ref{fig:environment} for an example). 

The appearance and positioning of the pieces is derived from symbolic piece representations: a tuple of shape (9), color (6), and position (8). We experiment with $360$ of these symbolic pieces which include all shapes, colors, and positions and split them into distinct sets. Therefore, the target symbols for the testing tasks are distinct from the ones seen during training (they might share color and shape though, but are for example positioned elsewhere). We ensure the reproducibility of our experiments by constructing $2500$ training, $175$ validation, and $420$ testing tasks representing scenes with a map size of $21\times21$ tiles (see Table~\ref{table:tasks} and Appendix~\ref{appendix:tasks} for the detailed generation process).

\section{The Follower Behaviors} %

For the Follower, we take inspiration from \citet{sun_adaplanner_2023} who suggest a plan-based approach towards solving text-based tasks with language models: given a task's natural language instruction the model initially produces a plan, which is then executed and repeatedly refined or revised. 
We implement a policy that keeps track of a plan that contains up to 10 actions (the plan horizon; which is exactly the number of actions needed to reach the diagonal corner of the partial view). Our Follower's behavior of following the plan is adjustable along two dimensions: confidence and autonomy.

\paragraph{Confidence.} The actions in the plan are associated with a decreasing probability of being executed (the ``confidence triangle'' in Figure~\ref{fig:environment}) so that given a discount factor $\phi \in [0,1]$ and a lower threshold $\mathrm{L} \in [0,1]$ we calculate:

\begin{equation}
    \text{Confidence}(a_i) = \text{max}(\phi^{i}, \mathrm{L})
\end{equation}

Which introduces a notion of confidence: either the planned action is executed or a wait action occurs (hesitation). Furthermore, this conceptualizes that a Follower becomes increasingly unsure about the continuation of the plan without receiving feedback from the Guide. 

\paragraph{Autonomy.} The revision process for our Follower policy is conceptually divided into five sub-programs that run after the Guide's utterance is received, parsed and the assumed intent type is determined, as follows:

\begin{itemize}
    \item \texttt{on\_silence}: The Follower executes, based on confidence, the next action in the plan (if available). Otherwise, it waits.
    \item \texttt{on\_confirm}: The Follower sets the confidence for all actions in the current plan to 1. Then the next action is chosen as described under \texttt{on\_silence}.
    \item \texttt{on\_decline}: The Follower erases the current plan. As the plan is then empty, a wait action will be returned.
    \item \texttt{on\_directive}: The Follower parses the utterances for the concrete directives (a direction or a ``take'' prompt). For ``take'', the plan is replaced with take action under the assumption that this is the last action to be performed. Otherwise, the plan is filled with actions that align with the direction prompt. Then, the next action is chosen as described under \texttt{on\_silence}.
    \item \texttt{on\_reference}: The Follower updates its internal target descriptor (color, shape, position) based on the new reference. Given this updated descriptor, the Follower identifies candidate coordinates in the symbolic representation of the current field of view, for example, coordinates that are blue given a reference ``Take the blue piece''. If such a coordinate is identified and the Follower has not already approached it, then the shortest path to that candidate is established as a new plan. Otherwise, if the descriptor only contains a position, then a direction towards that position is approached. In the case where the Follower is already in that position, a randomly chosen piece in the field of view is approached. When none of this matches, then the current plan proceeds as described under \texttt{on\_silence}.
\end{itemize}

Now, the autonomy defines which procedures the Follower undertakes, when intermediate feedback \textit{is missing} (the Guide stays silent). The \textbf{cautious} Follower is performing solely the previously defined procedures: when the plan is exhausted, then it waits until a new directive or reference is given. If this Follower is over an assumed target piece, then it waits until the ``take'' directive is given by the Guide. In contrast, the \textbf{eager} Follower aims to actually take an assumed target piece when approaching it in the current field of view. Furthermore, the eager Follower autonomously looks for target candidates at each step (as described in the \texttt{on\_reference} procedure) and potentially revises the plan (also when the Guide stays silent). 

\section{Learning Communication Policies for Different Follower Behaviors} %

\citet{mnih_human-level_2015} showed that vision-driven reinforcement learning policies can achieve human-level performance in pixel-based environments like Atari games. Similarly, the Guide as an agent in our environment has the challenging task to learn:
\begin{itemize}
    \item when to produce an utterance (or stay silent),
    \item what to produce (confirm, decline, direct, refer), and 
    \item how to produce it (which directive or preference order)
\end{itemize}
based solely on visual observation of the board state and the follower actions.

\begin{figure}[b]
    \begin{center}
        \includegraphics[width=0.45\textwidth]{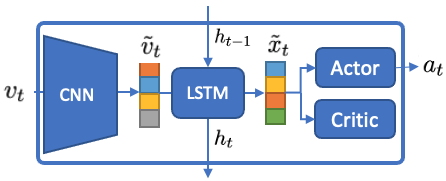}
    \end{center}
    \caption{The Guide's recurrent vision network.}
    \label{fig:guide}
\end{figure}

\subsection{The Guide} 

The observation $o_t=(v_t)$ with $v_t \in \mathbb{R}^{21 \times 21 \times 5}$ is encoded into a 128-dimensional feature vector $\tilde{v}_t \in \mathbb{R}$ using a 4-layer convolutional neural network similar to that by \citet{chevalier-boisvert_babyai_2019}. Then, the feature vector $\tilde{v}_t$ is fed through an LSTM \citep{hochreiter_long_1997} which functions as a memory mechanism (updating a state vector $h_t$ that is passed forward in time). Given the resulting memory-conditioned visual feature vector $\tilde{x}_t$, we learn a parameterized actor-critic-based policy $\pi(\tilde{x}_t;\theta) \sim a_t$ where the actor predicts a distribution over the action space (intents) and the critic estimates the value of the current state (Figure~\ref{fig:guide}). For the recurrent policy, %
we use the implementation of \textit{StableBaselines3-Contrib} v1.8.0 \citep{stable-baselines3}, which performs back-propagation through time until the first step in an episode.
 
\subsection{Experiment Setup} 

In this work, we evaluate if and to which extent the PPO algorithm~\citep{schulman_proximal_2017} is able to produce guiding neural agents in a challenging reference game where the learning signal is a sparse reward that also involves the assumed accumulated effort over actions. In particular, we are interested in the question of whether the resulting learned policies (the Guides) are adapted towards the Follower behaviors in such ways that align with expectations based on the Follower's dimensions of confidence and autonomy. Thus, for the experiments, we initiate \textbf{cautious} and \textbf{eager} Follower's with increasing confidence discount factors so that $\phi \in [0.75, 0.85, 0.90, 0.95, 0.97, 0.99]$.

We use \textit{StableBaselines3} v1.8.0~\citep{stable-baselines3} to train for each of these Follower behaviors a policy. We train each policy with 4 parallel running environments (batch size) and 1 million time steps in total. This means that each board in the training split is seen at least 13 times (and even more often when  mean $T < 30$). Every 100k steps during training, we evaluate the policies against the validation set. And we  keep the policies (the Guides) that achieve the highest mean episode reward based on these validation runs. We conduct the experiments with three different seeds.

\begin{table}
\small
\centering
\begin{tabular}{|rcccc|}
\hline
\multicolumn{1}{|r}{\textbf{Metrics:}}                            & \textbf{mR} $\uparrow$ & \textbf{mSR} $\uparrow$ & \textbf{mEPL} $\downarrow$ & \textbf{mEff.} $\downarrow$ \\ \hline
\multicolumn{5}{|c|}{\textbf{--- Cautious ---}}\\ 
\multicolumn{1}{|l}{\textbf{100\% Silent}}           & 0.00            & 0.00          & 30.00          & 0.00            \\
\multicolumn{1}{|l}{\textbf{100\% Ref.}}           & -1.04            & 0.00          & 30.00          & 34.8            \\ 
\multicolumn{1}{|l}{\textbf{PPO-Guide}}           & 1.55            & 0.94          & 13.97          & 10.72            \\ 
\rowcolor[HTML]{EFEFEF}
\multicolumn{1}{|r|}{$\phi$=75}                         & 1.52            & 0.93          & 15.02          & 11.07           \\ 
\multicolumn{1}{|r|}{$\phi$=85} & 1.47            & \textbf{0.96}          & 14.13          & 14.63            \\
\rowcolor[HTML]{EFEFEF}
\multicolumn{1}{|r|}{$\phi$=90}                         & 1.59            & 0.95          & 13.87          & 10.33            \\
\multicolumn{1}{|r|}{$\phi$=95} & 1.57            & 0.94          & 13.67          & 10.49           \\
\rowcolor[HTML]{EFEFEF}
\multicolumn{1}{|r|}{$\phi$=97}                         & 1.57            & 0.93          & 13.27          & 10.00            \\
\multicolumn{1}{|r|}{$\phi$=99} & 1.57            & 0.90          & 13.88          & 7.78            \\ \hline
\multicolumn{5}{|c|}{\textbf{--- Eager ---}}\\ 
\multicolumn{1}{|l}{\textbf{100\% Silent}}              & 0.45            & 0.23          & 16.78          & 0.00            \\
\multicolumn{1}{|l}{\textbf{100\% Ref.}}              & 0.86            & 0.75          & 18.57          & 21.09            \\
\multicolumn{1}{|l}{\textbf{PPO-Guide}}              & 1.57            & 0.91          & 13.19          & 8.72            \\ 
\rowcolor[HTML]{EFEFEF}
\multicolumn{1}{|r|}{$\phi$=75}                         & 1.54            & 0.92 & 13.54          & 10.04           \\
\multicolumn{1}{|r|}{$\phi$=85} & \textbf{1.60}   & 0.89          & 14.28          & \textbf{6.15}   \\
\rowcolor[HTML]{EFEFEF}
\multicolumn{1}{|r|}{$\phi$=90}                         & 1.49            & 0.92          & 13.24          & 11.67            \\ 
\multicolumn{1}{|r|}{$\phi$=95} & 1.59            & 0.92          & 12.86          & 8.39            \\
\rowcolor[HTML]{EFEFEF}
\multicolumn{1}{|r|}{$\phi$=97}                         & 1.58            & 0.90          & 12.64          & 7.28            \\ 
\multicolumn{1}{|r|}{$\phi$=99} & 1.59            & 0.93          & \textbf{12.58} & 8.76            \\ \hline
\multicolumn{5}{|c|}{\textbf{--- Overall ---}} \\
\multicolumn{1}{|l}{\textbf{\textbf{100\% Silent}}}            & 0.23            & 0.11          & 23.39          & 0.00   \\
\multicolumn{1}{|l}{\textbf{\textbf{100\% Ref.}}}            & -0.09            & 0.37          & 24.29          & 27.94   \\
\multicolumn{1}{|l}{\textbf{\textbf{PPO-Guide}}}            & 1.56            & 0.92          & 13.58          & 9.72  \\ \hline
\end{tabular}
\caption{The mean rewards (mR), success rates (mSR in \%), episodes lengths (mEPL) and efforts of the agents on the test tasks for the chosen autonomy and confidence combinations of the Follower (averaged over all seeds). A shortest path solver reaches $11.65$ mEPL ($3.13$ std). Given this, the upper bound for the mean reward is $1.83$. Best values in bold.}
\label{table:results_main}
\end{table}

\begin{table}
\centering
\small
\begin{tabular}{|rccccc|}
\hline
\multicolumn{1}{|r}{\textbf{Chosen Intent:}}                            & \multicolumn{1}{c}{\textbf{S}} & \multicolumn{1}{c}{\textbf{C}} & \multicolumn{1}{c}{\textbf{D}} & \multicolumn{1}{c}{\textbf{O}} & \textbf{R} \\ \hline
\multicolumn{6}{|c|}{\textbf{--- Cautious ---}} \\
\textbf{PPO-Guide}                                & 0.27                           & 0.04                           & /                              & 0.09                           & 0.60       \\ 
\rowcolor[HTML]{EFEFEF}
\multicolumn{1}{|r|}{$\phi$=75}                         & 0.27                           & 0.08                           & /                              & 0.08                           & 0.56       \\
\multicolumn{1}{|r|}{$\phi$=85} & 0.06                           & 0.08                              & /                              & 0.09                           & 0.78       \\
\rowcolor[HTML]{EFEFEF}
\multicolumn{1}{|r|}{$\phi$=90}                         & 0.29                           & 0.09                           & /                              & 0.08                           & 0.53       \\
\multicolumn{1}{|r|}{$\phi$=95} & 0.28                           & /                           & /                              & 0.09                           & 0.63       \\
\rowcolor[HTML]{EFEFEF}
\multicolumn{1}{|r|}{$\phi$=97}                         & 0.30                           & /                              & /                              & 0.09                           & 0.61       \\
\multicolumn{1}{|r|}{$\phi$=99} & 0.43                           & /                              & /                              & 0.09                           & 0.48       \\ \hline
\multicolumn{6}{|c|}{\textbf{--- Eager ---}} \\
\textbf{PPO-Guide}                                   & 0.34                           & 0.06                           & 0.06                           & 0.09                           & 0.46       \\
\rowcolor[HTML]{EFEFEF}
\multicolumn{1}{|r|}{$\phi$=75}                         & 0.25                           & 0.26                           & 0.03                           & 0.08                           & 0.38       \\
\multicolumn{1}{|r|}{$\phi$=85} & 0.53                           & 0.01                              & 0.09                           & 0.08                           & 0.29       \\
\rowcolor[HTML]{EFEFEF}
\multicolumn{1}{|r|}{$\phi$=90}                         & 0.16                           & 0.05                              & 0.11                           & 0.08                           & 0.59       \\
\multicolumn{1}{|r|}{$\phi$=95} & 0.34                           & /                              & 0.13                             & 0.09                           & 0.45       \\
\rowcolor[HTML]{EFEFEF}
\multicolumn{1}{|r|}{$\phi$=97}                         & 0.42                           & /                              & /                              & 0.11                           & 0.47       \\
\multicolumn{1}{|r|}{$\phi$=99} & 0.33                           & 0.02                           & /                           & 0.08                           & 0.57       \\ \hline
\multicolumn{6}{|c|}{\textbf{--- Overall ---}} \\
\textbf{PPO-Guide}                                 & 0.31                           & 0.05                           & 0.03                           & 0.09                           & 0.53       \\ \hline
\end{tabular}
\caption{The intent’s mean chance of being chosen at a step (for each policy evaluated on the test split) broken down by a Follower's confidence and autonomy. The intents are abbreviated as follows: \texttt{silence} (S), \texttt{confirm} (C), \texttt{decline} (D), \texttt{directive} (O) and \texttt{reference} (R). It appears reasonable that the cautious Follower's actions are never declined because the behavior is to always wait for the Guide's instructions (in contrast to the eager ones that explore occasionally on their own). Similarly, the higher confidence Follower's require less re-assurance (confirms) of their actions.}
\label{table:utterances}
\end{table}

\subsection{Results and Discussion} %

\paragraph{Overall Results.} The overall results in Table~\ref{table:results_main} show that learned  policies are communicative strategies that can successfully guide the Follower (towards the target piece) in most of the cases (on average in 92\% of the test episodes). This indicates that the Guide learned the goal of the game and hereby almost reaches the best episode length (on average only $1.93$ steps longer than the shortest path). The overall average effort (9.72) covers only about $71.5$\% of the average episode length (13.58) which means that the policies altogether produce an utterance in about 2 out of 3 steps.

\begin{figure}
    \begin{center}
        \includegraphics[width=0.41\textwidth]{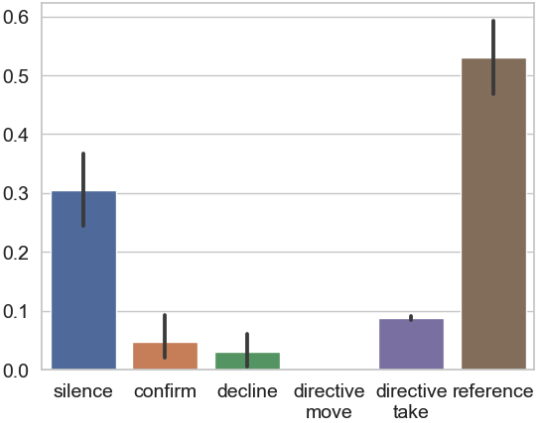}
    \end{center}
    \caption{An intent's mean chance of being chosen at a step (for all learnt policies evaluated on the test split).}
    \label{fig:intents}
\end{figure}

\paragraph{Has the Guide learned to stay silent?} Indeed, Figure~\ref{fig:intents} shows that the policies converge to a mode where the \texttt{silence} intent is chosen in at least 23\% of the steps: The policies are in general able to learn to say nothing. The most chosen intent is \texttt{reference}, which is reasonable as it is directly providing additional information (the target piece description) to the Follower and triggers a plan revision.

\begin{figure}
    \begin{center}
        \includegraphics[width=0.41\textwidth]{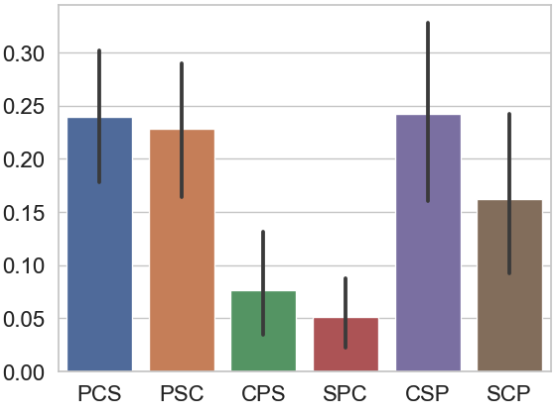}
    \end{center}
    \caption{The distribution of the preference order choices for the \texttt{reference} action (from Figure~\ref{fig:intents}). The preferences over position (P), shape (S) and color (C) are given to the \textsc{ia} for reference production.}
    \label{fig:preferences}
\end{figure}

\begin{figure}
    \begin{center}
        \includegraphics[width=0.41\textwidth]{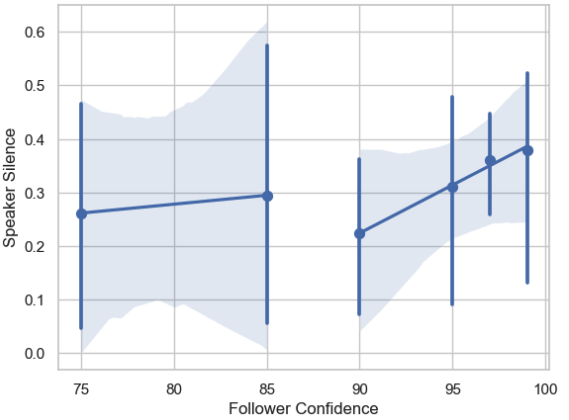}
    \end{center}
    \caption{The linear regressions with a confidence interval of 99\% for the mean silence rates measured during the test episodes for all learnt policies (3 seeds per follower). Fitted separately for the confidences $\{75,85\}$ and $\{90,95,97,99\}$.}
    \label{fig:regressions}
\end{figure}

\paragraph{What preference orders are chosen for the \texttt{reference} production?}

The \texttt{reference} intents define the order in which properties are compared between the target piece and its distractors. This means, for example, that a \textsc{csp} reference is likely to mention the target piece's color because the color attribute is first compared to distinguish the target from its distractors (and it is very likely that at least one distractor gets excluded, because otherwise all pieces would share the same color). Thus, it is reasonable that there are communicative strategies learnt that choose \textsc{csp} in the majority of cases. This means that the guide produces a reference that likely includes the shape and the color of the target piece. These properties are indeed useful for the follower to identify and approach the target in its field of view. An the other hand, preference orders that test positions first (\textsc{pcs} and \textsc{psc}) are also chosen rather often. These strategies lead the Follower to the target piece without having it necessarily already in the field of view.

\paragraph{The effects of the Follower's autonomy mode.}

We experimented with two levels of autonomy of the Follower. The results in Table~\ref{table:results_main} show that the policies that learn from interactions with the \textbf{eager} Follower require on average $2.00$ points less effort than the \textbf{cautious} one. This is reasonable as the eager Follower is autonomously updating the plan and looking for target candidates at each step. Along these lines, it is also reasonable that the decline intent is never selected for the cautious Follower (see Table~\ref{table:utterances}) because it never tried to approach a target piece without the Guide referencing it. 

\paragraph{The effects of the Follower's confidence.}

The differences in the intent selection strategy of the learned policies (Guides) shown in Table~\ref{table:utterances} indicate that Guides learnt from interaction with more confident Follower's ($\phi > 0.9)$ produce less or no \texttt{confirm} actions. This seems reasonable as the decrease in the execution probability of these Followers is less steep and a \texttt{reference} action has a similar effect. Furthermore, we see a slight tendency of Guide's to stay quieter (on average) when trained with more confident Followers as shown in Figure~\ref{fig:regressions}. Though we cannot see such a tendency for Guide's trained with less confident Followers.

\section{Conclusions} %

In this work, we examined an interesting intersection between psycho-linguistic studies and deep learning with reinforcement learning. We considered neural agents as possible interaction partners (for humans) in a challenging reference game where a Guide has to learn when, what, and how information (actionable intents) is to be provided to a Follower. As a proxy for different Follower behaviors, we implemented a hand-crafted policy that is controllable along two dimensions: autonomy in exploration and confidence in executing an action. We experimented with a learning signal that in addition to the goal condition also respects an assumed communicative effort. Our results indicate that this formulation of the learning signal leads to communicative strategies that are less verbose (stay silent more often) and that the resulting Guide behaviors are adapted (in terms of intent selection distributions) to the Follower's autonomy and confidence levels. We think this work presents a useful case study towards neural agents that have to learn adapted communication strategies in an interactive setting (possibly with humans). In future work, we want to investigate other reward formulations for our proven reference game and evaluate the learning of communication policies in an even more incremental setting where the utterance production process spans multiple time steps (one word at a time) and must be possibly interrupted and re-adjusted during the interaction.

\section*{Acknowledgements} We want to thank the anonymous reviewers for their comments. This work was funded by the \textit{Deutsche Forschungsgemeinschaft} (DFG, German Research Foundation) – 423217434 (``RECOLAGE'') grant.

\appendix

\section{Appendix}

Robot image in Figure~\ref{fig:example_board} adjusted from \url{https://commons.wikimedia.org/wiki/File:Cartoon_Robot.svg}.
That file was made available under the Creative Commons CC0 1.0 Universal Public Domain Dedication.

\subsection{Environment Details}
\label{appendix:environment}

\paragraph{Board} The internal representation of the visual state is a 2-dimensional grid that spans $W \times H$ tiles where $W$ and $H$ are defined by the map size. A tile is either empty or holds an identifier for a piece (the tile is then occupied). The pieces are defined by their colour, shape and coordinates and occupy five adjacent tiles (within a virtual box of $5\times5$ tiles). The pieces are not allowed to overlap with another piece's tiles. For a higher visual variation, we also apply rotations to pieces, but we ignore the rotation for expression generation, though this could be an extension of the task. The colors are described in Table~\ref{tab:colors}.

\begin{table}[h]
    \centering
    \begin{tabular}{|l|l|l|c|}
    \hline
    Name & HEX & RGB\\
    \hline
 red & \#ff0000 & (255, 0, 0)\\
 green & \#008000 & (0, 128, 0) \\
 blue & \#0000ff & (0, 0, 255)\\
 yellow & \#ffff00 & (255, 255, 0) \\
 brown & \#8b4513 & (139, 69, 19)\\
 purple & \#800080 & (128, 0, 128)\\
    \hline
    \end{tabular}
    \caption{The colors for the Pentomino pieces.}
    \label{tab:colors}
    \vspace{-0.3cm}
\end{table}

\paragraph{Symbols} The symbolic repesentations for the shapes are: P (2), X (3), T (4), Z (5), W (6), U (7), N (8), F (9), Y (10). The colors are encoded as: red (2), green (3), blue (4), yellow (5), brown (6), purple (7). The 0-symbol is reserved for out-of-world tiles (which can occur in the partial view). The 1-symbol is reserved for an empty tile.

\paragraph{Gripper} The gripper can only move one position at a step and can move over pieces, but is not allowed to leave the boundaries of the board. The gripper coordinates are defined as $\{(x,y): x \in [0,W], y \in [0,H]\}$.

\paragraph{References}

The Incremental Algorithm (Algorithm~\ref{alg:ia}), in the formulation of \cite{dale_computational_1995},
is supposed to find the properties that uniquely identify an object among others given a preference over properties. To accomplish this the algorithm is given the property values $\mathcal{P}$ of distractors in $M$ and of a referent $r$. Then the algorithm excludes distractors in several iterations until either $M$ is empty or every property of $r$ has been tested. During the exclusion process the algorithm computes the set of distractors that do \textit{not} share a given property with the referent and stores the property in $\mathcal{D}$. These properties in $\mathcal{D}$ are the ones that distinguish the referent from the others and thus will be returned.

The algorithm has a meta-parameter $\mathcal{O}$, indicating the \textit{preference order}, which determines the order in which the properties of the referent are tested against the distractors. In our domain, for example, when \textit{color} is the most preferred property, the algorithm might return \textsc{blue}, if this property already excludes all distractors. When \textit{shape} is the preferred property and all distractors do \textit{not} share the shape \textsc{T} with the referent, \textsc{T} would be returned. Hence even when the referent and distractor pieces are the same, different preference orders might lead to different expressions.

\begin{algorithm}[t]
\caption{The \textsc{ia} on symbolic properties as based on the formulation by \citet{deemter_2016}}\label{alg:ia}
\begin{algorithmic}[1]
\Require{A set of distractors $M$, a set of property values $\mathcal{P}$ of a referent $r$ and a linear preference order $\mathcal{O}$ over the property values $\mathcal{P}$}
\State{$\mathcal{D} \gets \emptyset $}
\For{$P$ in $\mathcal{O}(\mathcal{P})$}
{
\State{$\mathcal{E} \gets \{ m \in M: \neg P(m)$\}}
\If{$\mathcal{E} \ne \emptyset$}
    \State Add $P$ to $\mathcal{D}$
    \State Remove $\mathcal{E}$ from $M$
\EndIf
\EndFor
}
\State{\textbf{return} $\mathcal{D}$}
\end{algorithmic}
\end{algorithm}

There are 3 expression templates that are used when only a single property value of the target piece is returned by the Incremental Algorithm (\textsc{ia}): 
\begin{itemize}
    \itemsep0em
    \item \textit{Take the [color] piece}
    \item \textit{Take the [shape]}
    \item \textit{Take the piece at [position]}
\end{itemize}
Then there are 3 expression templates that are selected when two properties are returned:
\begin{itemize}
    \itemsep0em
    \item \textit{Take the [color] [shape]}
    \item \textit{Take the [color] piece at [position]}
    \item \textit{Take the [shape] at [position]}
\end{itemize}
And finally there is one expression templates that lists all property values to identify a target piece:
\begin{itemize}
    \itemsep0em
    \item \textit{Take the [color] [shape] at [position]}
\end{itemize}

\paragraph{Vocabulary}
\label{appendix:vocabulary}

 Overall, the property values and sentence templates lead to a small vocabulary of 37 words:

\begin{itemize}
    \item 9 shapes: P, X, T, Z, W, U, N, F, Y
    \item 6 colors: red, green, blue, yellow, brown, purple
    \item 6 position words: left, right, top, bottom, center (which are combined to e.g., right center or top left)
    \item 12 template words: take, the, piece, at, yes, no, this, way, go, a, bit, more
    \item 4 special words: \texttt{<s>}, \texttt{<e>}, \texttt{<pad>}, \texttt{<unk>}
\end{itemize}

The maximal sentence length is 12.

\subsection{Task Details}

\label{appendix:tasks}
To create a task, we first place the target piece on a board. Then, we sample uniformly random from all possible pieces and place them until the wanted number of pieces is reached (we experiment with 2 to 8 pieces on a board). If a piece cannot be placed after a certain amount of tries, then we re-sample a piece and try again. The coordinates are chosen at random uniform from the coordinates that fall into an area of the symbolic description. We never set a piece into the center, because that is the location where the gripper is initially located. 
In this way, we construct 100 training boards (or 1 evaluation board respectively) for each number of pieces (2-8). To ensure that a board scene in the training split cannot be aligned with a target piece, we create 3 extra tasks for a single board by choosing extra targets (when fewer than 4 pieces are on a board, then we create a task for each piece). For evaluation, we only create a single task for each target piece symbol.

\subsection{Guide Details}
\label{appendix:guide}

\paragraph{Agent}
Parameters: $602,447$

\begin{table}[h]
    \centering
    \begin{tabular}{| l | r| }
        \hline
         feature\_dims & 128 \\
         normalize\_images & True \\
         shared\_lstm & True \\
         enable\_critic\_lstm & False \\
         n\_lstm\_layers & 1 \\
         lstm\_hidden\_size & 128 \\
        \hline
    \end{tabular}
    \caption{Policy arguments for the the RecurrentPPO agent}
    \label{tab:agent_hyperparameters}
\end{table}

\paragraph{Policy Architecture} We instantiate the actor-critic PPO agent with an architecture defined by \texttt{pi=[64, 64], vf=[64, 64]} meaning that the actor is a 2-layer feedforward network with 64 parameters per layer. The critic has the same architecture, but does not share the weights with the actor.

\paragraph{Vision Encoder} The visual encoder is a convolutional neural network (CNN) with 4 layers that maps the visual observations $v_t \in \mathbb{R}^{21\times21\times5}$ into a 128-dimensional features vector $\tilde{v} \in \mathbb{R}$. We consecutively apply four blocks of \texttt{(nn.Conv2d(),nn.BatchNorm2d(),nn.ReLU())} with same padding where the kernel size is $3\times3$, except for the first blocke where we set the kernel size to $1\times1$. After the fourth block we apply a \texttt{nn.AdaptiveMaxPool2d((1, 1))} layer from PyTorch v1.13.0 \cite{DBLP:conf/nips/PaszkeGMLBCKLGA19} to collapse the spatial dimensions of the feature maps.

\paragraph{Learning Algorithm}

We use the RecurrentPPO implementation from StableBaselines-Contrib v1.8.0 \citep{stable-baselines3} with the hyper-parameters in Table~\ref{tab:ppo_hyperparameters} (and the defaults otherwise).

\begin{table}[h]
    \centering
    \begin{tabular}{| l | r| }
        \hline
        learning\_rate & 3e-4 \\
         clip\_range & 0.2 \\
         gamma & 0.99 \\
         gae\_lambda & 0.95\\
         ent\_coef & 0.0 \\
         vf\_coef & 0.5 \\
         max\_grad\_norm & 0.5 \\
         lr\_init & 3e-4 \\
         n\_steps & 128 \\
         batch\_size & 128 \\
         num\_epochs & 10\\
        \hline
    \end{tabular}
    \caption{RecurrentPPO hyperparameters}
    \label{tab:ppo_hyperparameters}
\end{table}

\subsection{Experiment Details}
\label{appendix:experiment}

We trained the agents simultaneously on 8 GeForce GTX 1080 Ti (11GB) where each of them consumed about 4GB of GPU memory. The training for the $36$ configurations took around $144$ hours in total (about $4h$ for the 1 million steps each). The random seeds were set to $49184$, $98506$ or $92999$ respectively.
As the evaluation criteria on the testings tasks we chose success rate which indicates the relative number of episodes (in a rollout or in a test split) where the agent selected the correct piece: 
$$
\text{mSR}=\frac{\sum^{N}{s_i}}{N} \text{ where } s_i=\begin{cases}
    1, & \text{for correct piece} \\
    0, & \text{otherwise}
\end{cases}
$$

\bibliography{aaai24}

\begin{thebibliography}{48}
\providecommand{\natexlab}[1]{#1}

\bibitem[{Albrecht and Stone(2018)}]{albrecht_autonomous_2018}
Albrecht, S.~V.; and Stone, P. 2018.
\newblock Autonomous agents modelling other agents: {A} comprehensive survey and open problems.
\newblock \emph{Artif. Intell.}, 258: 66--95.

\bibitem[{Chane{-}Sane, Schmid, and Laptev(2021)}]{chane-sane_goal-conditioned_2021}
Chane{-}Sane, E.; Schmid, C.; and Laptev, I. 2021.
\newblock Goal-Conditioned Reinforcement Learning with Imagined Subgoals.
\newblock In Meila, M.; and Zhang, T., eds., \emph{Proceedings of the 38th International Conference on Machine Learning, {ICML} 2021, 18-24 July 2021, Virtual Event}, volume 139 of \emph{Proceedings of Machine Learning Research}, 1430--1440. {PMLR}.

\bibitem[{Chang et~al.(2022)Chang, Mu, Wu, Pan, and Xu}]{chang_e-mapp_nodate}
Chang, C.; Mu, N.; Wu, J.; Pan, L.; and Xu, H. 2022.
\newblock {E-MAPP:} Efficient Multi-Agent Reinforcement Learning with Parallel Program Guidance.
\newblock In \emph{NeurIPS}.

\bibitem[{Chevalier{-}Boisvert et~al.(2019)Chevalier{-}Boisvert, Bahdanau, Lahlou, Willems, Saharia, Nguyen, and Bengio}]{chevalier-boisvert_babyai_2019}
Chevalier{-}Boisvert, M.; Bahdanau, D.; Lahlou, S.; Willems, L.; Saharia, C.; Nguyen, T.~H.; and Bengio, Y. 2019.
\newblock BabyAI: {A} Platform to Study the Sample Efficiency of Grounded Language Learning.
\newblock In \emph{7th International Conference on Learning Representations, {ICLR} 2019, New Orleans, LA, USA, May 6-9, 2019}. OpenReview.net.

\bibitem[{Clark(1996)}]{clark_using_1996}
Clark, H.~H. 1996.
\newblock \emph{Using {Language}}.
\newblock '{Using}' {Linguistic} {Books}. Cambridge: Cambridge University Press.
\newblock ISBN 978-0-521-56158-7.

\bibitem[{Clark and Wilkes-Gibbs(1986)}]{clark_referring_1986}
Clark, H.~H.; and Wilkes-Gibbs, D. 1986.
\newblock Referring as a collaborative process.
\newblock \emph{Cognition}, 22(1): 1--39.
\newblock Place: Netherlands Publisher: Elsevier Science.

\bibitem[{Co{-}Reyes et~al.(2019)Co{-}Reyes, Gupta, Sanjeev, Altieri, Andreas, DeNero, Abbeel, and Levine}]{co-reyes_guiding_2019}
Co{-}Reyes, J.~D.; Gupta, A.; Sanjeev, S.; Altieri, N.; Andreas, J.; DeNero, J.; Abbeel, P.; and Levine, S. 2019.
\newblock Guiding Policies with Language via Meta-Learning.
\newblock In \emph{7th International Conference on Learning Representations, {ICLR} 2019, New Orleans, LA, USA, May 6-9, 2019}. OpenReview.net.

\bibitem[{Dale and Reiter(1995)}]{dale_computational_1995}
Dale, R.; and Reiter, E. 1995.
\newblock Computational Interpretations of the Gricean Maxims in the Generation of Referring Expressions.
\newblock \emph{Cogn. Sci.}, 19(2): 233--263.

\bibitem[{Davidesco et~al.(2023)Davidesco, Laurent, Valk, West, Milne, Poeppel, and Dikker}]{davidesco_temporal_2023}
Davidesco, I.; Laurent, E.; Valk, H.; West, T.; Milne, C.; Poeppel, D.; and Dikker, S. 2023.
\newblock The {Temporal} {Dynamics} of {Brain}-to-{Brain} {Synchrony} {Between} {Students} and {Teachers} {Predict} {Learning} {Outcomes}.
\newblock \emph{Psychological Science}, 34(5): 633--643.

\bibitem[{E et~al.(2019)E, Niu, Chen, and Song}]{e_novel_2019}
E, H.; Niu, P.; Chen, Z.; and Song, M. 2019.
\newblock A Novel Bi-directional Interrelated Model for Joint Intent Detection and Slot Filling.
\newblock In Korhonen, A.; Traum, D.~R.; and M{\`{a}}rquez, L., eds., \emph{Proceedings of the 57th Conference of the Association for Computational Linguistics, {ACL} 2019, Florence, Italy, July 28- August 2, 2019, Volume 1: Long Papers}, 5467--5471. Association for Computational Linguistics.

\bibitem[{Fried et~al.(2018)Fried, Hu, Cirik, Rohrbach, Andreas, Morency, Berg{-}Kirkpatrick, Saenko, Klein, and Darrell}]{DBLP:conf/nips/FriedHCRAMBSKD18}
Fried, D.; Hu, R.; Cirik, V.; Rohrbach, A.; Andreas, J.; Morency, L.; Berg{-}Kirkpatrick, T.; Saenko, K.; Klein, D.; and Darrell, T. 2018.
\newblock Speaker-Follower Models for Vision-and-Language Navigation.
\newblock In Bengio, S.; Wallach, H.~M.; Larochelle, H.; Grauman, K.; Cesa{-}Bianchi, N.; and Garnett, R., eds., \emph{Advances in Neural Information Processing Systems 31: Annual Conference on Neural Information Processing Systems 2018, NeurIPS 2018, December 3-8, 2018, Montr{\'{e}}al, Canada}, 3318--3329.

\bibitem[{Gao et~al.(2022)Gao, Gao, Gong, Lin, Thattai, and Sukhatme}]{gao_dialfred_2022}
Gao, X.; Gao, Q.; Gong, R.; Lin, K.; Thattai, G.; and Sukhatme, G.~S. 2022.
\newblock DialFRED: Dialogue-Enabled Agents for Embodied Instruction Following.
\newblock \emph{{IEEE} Robotics Autom. Lett.}, 7(4): 10049--10056.

\bibitem[{Ghosh et~al.(2020)Ghosh, Tschiatschek, Mahdavi, and Singla}]{ghosh_towards_2020}
Ghosh, A.; Tschiatschek, S.; Mahdavi, H.; and Singla, A. 2020.
\newblock Towards Deployment of Robust Cooperative {AI} Agents: An Algorithmic Framework for Learning Adaptive Policies.
\newblock In Seghrouchni, A. E.~F.; Sukthankar, G.; An, B.; and Yorke{-}Smith, N., eds., \emph{Proceedings of the 19th International Conference on Autonomous Agents and Multiagent Systems, {AAMAS} '20, Auckland, New Zealand, May 9-13, 2020}, 447--455. International Foundation for Autonomous Agents and Multiagent Systems.

\bibitem[{Golomb(1996)}]{golomb_1996}
Golomb, S.~W. 1996.
\newblock \emph{Polyominoes: Puzzles, Patterns, Problems, and Packings}.
\newblock Princeton University Press.
\newblock ISBN 0691024448.

\bibitem[{Gu et~al.(2022)Gu, Stefani, Wu, Thomason, and Wang}]{gu_vision-and-language_2022}
Gu, J.; Stefani, E.; Wu, Q.; Thomason, J.; and Wang, X. 2022.
\newblock Vision-and-Language Navigation: {A} Survey of Tasks, Methods, and Future Directions.
\newblock In Muresan, S.; Nakov, P.; and Villavicencio, A., eds., \emph{Proceedings of the 60th Annual Meeting of the Association for Computational Linguistics (Volume 1: Long Papers), {ACL} 2022, Dublin, Ireland, May 22-27, 2022}, 7606--7623. Association for Computational Linguistics.

\bibitem[{G{\"{u}}rtler, B{\"{u}}chler, and Martius(2021)}]{gurtler_hierarchical_2021}
G{\"{u}}rtler, N.; B{\"{u}}chler, D.; and Martius, G. 2021.
\newblock Hierarchical Reinforcement Learning with Timed Subgoals.
\newblock In Ranzato, M.; Beygelzimer, A.; Dauphin, Y.~N.; Liang, P.; and Vaughan, J.~W., eds., \emph{Advances in Neural Information Processing Systems 34: Annual Conference on Neural Information Processing Systems 2021, NeurIPS 2021, December 6-14, 2021, virtual}, 21732--21743.

\bibitem[{Hochreiter and Schmidhuber(1997)}]{hochreiter_long_1997}
Hochreiter, S.; and Schmidhuber, J. 1997.
\newblock Long {Short}-{Term} {Memory}.
\newblock \emph{Neural Computation}, 9(8): 1735--1780.

\bibitem[{Hu et~al.(2023)Hu, Zhao, Zhang, Zhou, Yang, Xu, and Liu}]{hu_enabling_2023}
Hu, B.; Zhao, C.; Zhang, P.; Zhou, Z.; Yang, Y.; Xu, Z.; and Liu, B. 2023.
\newblock Enabling Intelligent Interactions between an Agent and an {LLM:} {A} Reinforcement Learning Approach.
\newblock \emph{CoRR}, abs/2306.03604.

\bibitem[{Huang, Lipovetzky, and Cohn(2023)}]{huang_reminder_2023}
Huang, S.; Lipovetzky, N.; and Cohn, T. 2023.
\newblock A Reminder of its Brittleness: Language Reward Shaping May Hinder Learning for Instruction Following Agents.
\newblock \emph{CoRR}, abs/2305.16621.

\bibitem[{Jeon et~al.(2022)Jeon, Kim, Jung, and Sung}]{jeon_maser_2022}
Jeon, J.; Kim, W.; Jung, W.; and Sung, Y. 2022.
\newblock {MASER:} Multi-Agent Reinforcement Learning with Subgoals Generated from Experience Replay Buffer.
\newblock In Chaudhuri, K.; Jegelka, S.; Song, L.; Szepesv{\'{a}}ri, C.; Niu, G.; and Sabato, S., eds., \emph{International Conference on Machine Learning, {ICML} 2022, 17-23 July 2022, Baltimore, Maryland, {USA}}, volume 162 of \emph{Proceedings of Machine Learning Research}, 10041--10052. {PMLR}.

\bibitem[{Jurgenson and Tamar(2023)}]{jurgenson_goal-conditioned_2023}
Jurgenson, T.; and Tamar, A. 2023.
\newblock Goal-Conditioned Supervised Learning with Sub-Goal Prediction.
\newblock \emph{CoRR}, abs/2305.10171.

\bibitem[{Kolb et~al.(2019)Kolb, Lang, Bartsch, Gansekoele, Koopmanschap, Romor, Speck, Mul, and Bruni}]{kolb_learning_2019}
Kolb, B.; Lang, L.; Bartsch, H.; Gansekoele, A.; Koopmanschap, R.; Romor, L.; Speck, D.; Mul, M.; and Bruni, E. 2019.
\newblock Learning to Request Guidance in Emergent Communication.
\newblock \emph{CoRR}, abs/1912.05525.

\bibitem[{Lee and Kim(2023)}]{lee_controllable_2023}
Lee, G.~T.; and Kim, K.~J. 2023.
\newblock A Controllable Agent by Subgoals in Path Planning Using Goal-Conditioned Reinforcement Learning.
\newblock \emph{{IEEE} Access}, 11: 33812--33825.

\bibitem[{Liu et~al.(2022)Liu, Li, Xu, Dou, and Liu}]{liu_heterogeneous_nodate}
Liu, Y.; Li, Y.; Xu, X.; Dou, Y.; and Liu, D. 2022.
\newblock Heterogeneous Skill Learning for Multi-agent Tasks.
\newblock In \emph{NeurIPS}.

\bibitem[{Lowe et~al.(2019)Lowe, Foerster, Boureau, Pineau, and Dauphin}]{lowe_pitfalls_2019}
Lowe, R.; Foerster, J.~N.; Boureau, Y.; Pineau, J.; and Dauphin, Y.~N. 2019.
\newblock On the Pitfalls of Measuring Emergent Communication.
\newblock In Elkind, E.; Veloso, M.; Agmon, N.; and Taylor, M.~E., eds., \emph{Proceedings of the 18th International Conference on Autonomous Agents and MultiAgent Systems, {AAMAS} '19, Montreal, QC, Canada, May 13-17, 2019}, 693--701. International Foundation for Autonomous Agents and Multiagent Systems.

\bibitem[{Mnih et~al.(2015)Mnih, Kavukcuoglu, Silver, Rusu, Veness, Bellemare, Graves, Riedmiller, Fidjeland, Ostrovski, Petersen, Beattie, Sadik, Antonoglou, King, Kumaran, Wierstra, Legg, and Hassabis}]{mnih_human-level_2015}
Mnih, V.; Kavukcuoglu, K.; Silver, D.; Rusu, A.~A.; Veness, J.; Bellemare, M.~G.; Graves, A.; Riedmiller, M.~A.; Fidjeland, A.; Ostrovski, G.; Petersen, S.; Beattie, C.; Sadik, A.; Antonoglou, I.; King, H.; Kumaran, D.; Wierstra, D.; Legg, S.; and Hassabis, D. 2015.
\newblock Human-level control through deep reinforcement learning.
\newblock \emph{Nat.}, 518(7540): 529--533.

\bibitem[{Moon et~al.(2020)Moon, Kottur, Crook, De, Poddar, Levin, Whitney, Difranco, Beirami, Cho, Subba, and Geramifard}]{moon_situated_2020}
Moon, S.; Kottur, S.; Crook, P.~A.; De, A.; Poddar, S.; Levin, T.; Whitney, D.; Difranco, D.; Beirami, A.; Cho, E.; Subba, R.; and Geramifard, A. 2020.
\newblock Situated and Interactive Multimodal Conversations.
\newblock In Scott, D.; Bel, N.; and Zong, C., eds., \emph{Proceedings of the 28th International Conference on Computational Linguistics, {COLING} 2020, Barcelona, Spain (Online), December 8-13, 2020}, 1103--1121. International Committee on Computational Linguistics.

\bibitem[{Mordatch and Abbeel(2018)}]{mordatch2017emergence}
Mordatch, I.; and Abbeel, P. 2018.
\newblock Emergence of Grounded Compositional Language in Multi-Agent Populations.
\newblock In McIlraith, S.~A.; and Weinberger, K.~Q., eds., \emph{Proceedings of the Thirty-Second {AAAI} Conference on Artificial Intelligence, (AAAI-18), the 30th innovative Applications of Artificial Intelligence (IAAI-18), and the 8th {AAAI} Symposium on Educational Advances in Artificial Intelligence (EAAI-18), New Orleans, Louisiana, USA, February 2-7, 2018}, 1495--1502. {AAAI} Press.

\bibitem[{Mu et~al.(2022)Mu, Zhong, Raileanu, Jiang, Goodman, Rockt{\"{a}}schel, and Grefenstette}]{mu_improving_2022}
Mu, J.; Zhong, V.; Raileanu, R.; Jiang, M.; Goodman, N.~D.; Rockt{\"{a}}schel, T.; and Grefenstette, E. 2022.
\newblock Improving Intrinsic Exploration with Language Abstractions.
\newblock \emph{CoRR}, abs/2202.08938.

\bibitem[{Mul, Bouchacourt, and Bruni(2019)}]{mul_mastering_2019}
Mul, M.; Bouchacourt, D.; and Bruni, E. 2019.
\newblock Mastering emergent language: learning to guide in simulated navigation.
\newblock \emph{CoRR}, abs/1908.05135.

\bibitem[{Nguyen et~al.(2019)Nguyen, Dey, Brockett, and Dolan}]{DBLP:conf/cvpr/NguyenDBD19}
Nguyen, K.; Dey, D.; Brockett, C.; and Dolan, B. 2019.
\newblock Vision-Based Navigation With Language-Based Assistance via Imitation Learning With Indirect Intervention.
\newblock In \emph{{IEEE} Conference on Computer Vision and Pattern Recognition, {CVPR} 2019, Long Beach, CA, USA, June 16-20, 2019}, 12527--12537. Computer Vision Foundation / {IEEE}.

\bibitem[{Nguyen and III(2019)}]{DBLP:conf/emnlp/NguyenD19}
Nguyen, K.; and III, H.~D. 2019.
\newblock Help, Anna! Visual Navigation with Natural Multimodal Assistance via Retrospective Curiosity-Encouraging Imitation Learning.
\newblock In Inui, K.; Jiang, J.; Ng, V.; and Wan, X., eds., \emph{Proceedings of the 2019 Conference on Empirical Methods in Natural Language Processing and the 9th International Joint Conference on Natural Language Processing, {EMNLP-IJCNLP} 2019, Hong Kong, China, November 3-7, 2019}, 684--695. Association for Computational Linguistics.

\bibitem[{Padmakumar et~al.(2022)Padmakumar, Thomason, Shrivastava, Lange, Narayan{-}Chen, Gella, Piramuthu, T{\"{u}}r, and Hakkani{-}T{\"{u}}r}]{padmakumar_teach_2022}
Padmakumar, A.; Thomason, J.; Shrivastava, A.; Lange, P.; Narayan{-}Chen, A.; Gella, S.; Piramuthu, R.; T{\"{u}}r, G.; and Hakkani{-}T{\"{u}}r, D. 2022.
\newblock TEACh: Task-Driven Embodied Agents That Chat.
\newblock In \emph{Thirty-Sixth {AAAI} Conference on Artificial Intelligence, {AAAI} 2022, Thirty-Fourth Conference on Innovative Applications of Artificial Intelligence, {IAAI} 2022, The Twelveth Symposium on Educational Advances in Artificial Intelligence, {EAAI} 2022 Virtual Event, February 22 - March 1, 2022}, 2017--2025. {AAAI} Press.

\bibitem[{Park et~al.(2023)Park, O'Brien, Cai, Morris, Liang, and Bernstein}]{park2023generative}
Park, J.~S.; O'Brien, J.~C.; Cai, C.~J.; Morris, M.~R.; Liang, P.; and Bernstein, M.~S. 2023.
\newblock Generative Agents: Interactive Simulacra of Human Behavior.
\newblock \emph{CoRR}, abs/2304.03442.

\bibitem[{Pashevich, Schmid, and Sun(2021)}]{pashevich_episodic_2021}
Pashevich, A.; Schmid, C.; and Sun, C. 2021.
\newblock Episodic Transformer for Vision-and-Language Navigation.
\newblock In \emph{2021 {IEEE/CVF} International Conference on Computer Vision, {ICCV} 2021, Montreal, QC, Canada, October 10-17, 2021}, 15922--15932. {IEEE}.

\bibitem[{Paszke et~al.(2019)Paszke, Gross, Massa, Lerer, Bradbury, Chanan, Killeen, Lin, Gimelshein, Antiga, Desmaison, K{\"{o}}pf, Yang, DeVito, Raison, Tejani, Chilamkurthy, Steiner, Fang, Bai, and Chintala}]{DBLP:conf/nips/PaszkeGMLBCKLGA19}
Paszke, A.; Gross, S.; Massa, F.; Lerer, A.; Bradbury, J.; Chanan, G.; Killeen, T.; Lin, Z.; Gimelshein, N.; Antiga, L.; Desmaison, A.; K{\"{o}}pf, A.; Yang, E.~Z.; DeVito, Z.; Raison, M.; Tejani, A.; Chilamkurthy, S.; Steiner, B.; Fang, L.; Bai, J.; and Chintala, S. 2019.
\newblock PyTorch: An Imperative Style, High-Performance Deep Learning Library.
\newblock In Wallach, H.~M.; Larochelle, H.; Beygelzimer, A.; d'Alch{\'{e}}{-}Buc, F.; Fox, E.~B.; and Garnett, R., eds., \emph{Advances in Neural Information Processing Systems 32: Annual Conference on Neural Information Processing Systems 2019, NeurIPS 2019, December 8-14, 2019, Vancouver, BC, Canada}, 8024--8035.

\bibitem[{Pertsch et~al.(2020)Pertsch, Rybkin, Ebert, Zhou, Jayaraman, Finn, and Levine}]{pertsch_long-horizon_2020}
Pertsch, K.; Rybkin, O.; Ebert, F.; Zhou, S.; Jayaraman, D.; Finn, C.; and Levine, S. 2020.
\newblock Long-Horizon Visual Planning with Goal-Conditioned Hierarchical Predictors.
\newblock In Larochelle, H.; Ranzato, M.; Hadsell, R.; Balcan, M.; and Lin, H., eds., \emph{Advances in Neural Information Processing Systems 33: Annual Conference on Neural Information Processing Systems 2020, NeurIPS 2020, December 6-12, 2020, virtual}.

\bibitem[{Raffin et~al.(2021)Raffin, Hill, Gleave, Kanervisto, Ernestus, and Dormann}]{stable-baselines3}
Raffin, A.; Hill, A.; Gleave, A.; Kanervisto, A.; Ernestus, M.; and Dormann, N. 2021.
\newblock Stable-Baselines3: Reliable Reinforcement Learning Implementations.
\newblock \emph{Journal of Machine Learning Research}, 22(268): 1--8.

\bibitem[{Schulman et~al.(2017)Schulman, Wolski, Dhariwal, Radford, and Klimov}]{schulman_proximal_2017}
Schulman, J.; Wolski, F.; Dhariwal, P.; Radford, A.; and Klimov, O. 2017.
\newblock Proximal Policy Optimization Algorithms.
\newblock \emph{CoRR}, abs/1707.06347.

\bibitem[{Suhr and Artzi(2022)}]{suhr_continual_2022}
Suhr, A.; and Artzi, Y. 2022.
\newblock Continual Learning for Instruction Following from Realtime Feedback.
\newblock \emph{CoRR}, abs/2212.09710.

\bibitem[{Sun et~al.(2023)Sun, Zhuang, Kong, Dai, and Zhang}]{sun_adaplanner_2023}
Sun, H.; Zhuang, Y.; Kong, L.; Dai, B.; and Zhang, C. 2023.
\newblock AdaPlanner: Adaptive Planning from Feedback with Language Models.
\newblock \emph{CoRR}, abs/2305.16653.

\bibitem[{Sutton and Barto(2018)}]{Sutton1998}
Sutton, R.~S.; and Barto, A.~G. 2018.
\newblock \emph{Reinforcement Learning: An Introduction}.
\newblock The MIT Press, second edition.

\bibitem[{Thomason et~al.(2019)Thomason, Murray, Cakmak, and Zettlemoyer}]{DBLP:journals/corr/abs-1907-04957}
Thomason, J.; Murray, M.; Cakmak, M.; and Zettlemoyer, L. 2019.
\newblock Vision-and-Dialog Navigation.
\newblock \emph{CoRR}, abs/1907.04957.

\bibitem[{van Deemter(2016)}]{deemter_2016}
van Deemter, K. 2016.
\newblock \emph{Computational Models of Referring}, chapter 4.6.
\newblock The MIT Press.
\newblock ISBN 9780262034555.

\bibitem[{Wang et~al.(2021)Wang, Shih, Xie, and Sadigh}]{wang_influencing_2022}
Wang, W.~Z.; Shih, A.; Xie, A.; and Sadigh, D. 2021.
\newblock Influencing Towards Stable Multi-Agent Interactions.
\newblock In Faust, A.; Hsu, D.; and Neumann, G., eds., \emph{Conference on Robot Learning, 8-11 November 2021, London, {UK}}, volume 164 of \emph{Proceedings of Machine Learning Research}, 1132--1143. {PMLR}.

\bibitem[{Xie et~al.(2020)Xie, Losey, Tolsma, Finn, and Sadigh}]{xie_learning_2021}
Xie, A.; Losey, D.~P.; Tolsma, R.; Finn, C.; and Sadigh, D. 2020.
\newblock Learning Latent Representations to Influence Multi-Agent Interaction.
\newblock In Kober, J.; Ramos, F.; and Tomlin, C.~J., eds., \emph{4th Conference on Robot Learning, CoRL 2020, 16-18 November 2020, Virtual Event / Cambridge, MA, {USA}}, volume 155 of \emph{Proceedings of Machine Learning Research}, 575--588. {PMLR}.

\bibitem[{Yang et~al.(2022)Yang, Zhao, Hu, Zhou, Zhu, and Li}]{DBLP:conf/nips/Yang0HZZL22}
Yang, M.; Zhao, J.; Hu, X.; Zhou, W.; Zhu, J.; and Li, H. 2022.
\newblock {LDSA:} Learning Dynamic Subtask Assignment in Cooperative Multi-Agent Reinforcement Learning.
\newblock In \emph{NeurIPS}.

\bibitem[{Zarrieß et~al.(2016)Zarrieß, Hough, Kennington, Manuvinakurike, DeVault, Fernández, and Schlangen}]{zarries_pentoref_2016}
Zarrieß, S.; Hough, J.; Kennington, C.; Manuvinakurike, R.; DeVault, D.; Fernández, R.; and Schlangen, D. 2016.
\newblock {PentoRef}: {A} {Corpus} of {Spoken} {References} in {Task}-oriented {Dialogues}.
\newblock In \emph{Proceedings of the {Tenth} {International} {Conference} on {Language} {Resources} and {Evaluation} ({LREC}'16)}, 125--131. Portorož, Slovenia: European Language Resources Association (ELRA).

\end{thebibliography}

\end{document}